\documentclass{article}

\usepackage[preprint]{neurips_2022}

\usepackage[utf8]{inputenc} % allow utf-8 input
\usepackage[T1]{fontenc}    % use 8-bit T1 fonts
\usepackage[hidelinks]{hyperref}       % hyperlinks
\usepackage{url}            % simple URL typesetting
\usepackage{booktabs}       % professional-quality tables
\usepackage{amsfonts}       % blackboard math symbols
\usepackage{nicefrac}       % compact symbols for 1/2, etc.
\usepackage{microtype}      % microtypography
\usepackage{enumitem}

\usepackage{wrapfig}
\usepackage{mathtools}

\usepackage[usenames,dvipsnames,svgnames,table]{xcolor}  % allows better color names
\usepackage{soul}
\usepackage{todonotes}

\newcommand{\cutforspace}[1]{}

\newcommand{\codefont}{\fontfamily{lmtt}\selectfont}
\usepackage{listings}

\usepackage{parcolumns}
\lstdefinestyle{datalogstyle}{
	basicstyle={\codefont\small},  % the default font size is HUGE
	xleftmargin={6pt},
        xrightmargin={6pt},
        breakindent=0pt,
	frame=tb,
	stepnumber=1,
	firstnumber=1,
	numberfirstline=true,
	tabsize=2,
	extendedchars=true,
	breaklines=true,
	columns=fullflexible,
	keepspaces=true,
	escapeinside={@}{@},
	firstnumber=last,
	captionpos=b,
	commentstyle=\color{black!65},
	numberstyle=\tiny\color{black!65},
	stringstyle=\color{codepurple},
	breakatwhitespace=false, 
	keepspaces=true,                 
	numbersep=5pt,                  
	showspaces=false,                
	showstringspaces=false,
	showtabs=false,
	aboveskip={0.8\baselineskip},
	belowskip={0.2\baselineskip},
}
\usepackage{tikz}
\usetikzlibrary{shapes,calc,positioning}

\global

\usepackage[most]{tcolorbox}
\tcbset{
  aibox/.style={
    top=10pt,
    colback=white,
    colframe=black,
    boxrule=0.5pt,
    colbacktitle=black,
    enhanced,
    center,
    attach boxed title to top left={yshift=-0.1in,xshift=0.15in},
    boxed title style={boxrule=0pt,colframe=white,},
  }
}
\newtcolorbox{AIbox}[2][]{aibox,title=#2,#1}

\definecolor{aigold}{RGB}{244,210, 1} 
\definecolor{aigreen}{RGB}{213, 245, 227}

\definecolor{humanpurple}{RGB}{235, 222, 240} 
\definecolor{commentgray}{RGB}{86, 101, 115}

\definecolor{aired}{RGB}{255,180,181}

\usepackage[noabbrev,capitalize]{cleveref} % to standardize references
\crefname{equation}{equation}{equations}   % "equation" is lowercased, overriding capitalize option
\crefname{section}{section}{sections}      % "section" is lowercased, overriding capitalize option
\crefname{footnote}{footnote}{footnotes}   
\crefname{listing}{Chat}{Chats}
\crefname{assumption}{assumption}{assumptions}
\crefname{line}{line}{lines}   % for Datalog lines of code, use "rule" instead of "line"
\title{Can Large Language Models Play Text Games Well?\\Current State-of-the-Art and Open Questions}

\author{
 Chen Feng Tsai\thanks{\ Equal contribution. Work done during Chen Feng's and Sierra's internship at TTI-Chicago.}$\,\,^{1,6}$\ \ Xiaochen Zhou$^{*2,3}$\ \ Sierra S. Liu$^{4,6}$\ \  Jing Li$^{3}$\ \ Mo Yu$^{4}$\ \ Hongyuan Mei$^{6}$ \\
$^1$University of Chicago\quad
$^2$Syracuse University\quad
$^3$New Jersey Institute of Technology\\
$^4$Millburn High School\quad 
$^5$WeChat AI\quad 
$^6$Toyota Technological Institute at Chicago\\
\texttt{sierra.sijia.liu@gmail.com,\ jingli@njit.edu}\\
\texttt{moyumyu@global.tencent.com,\ hongyuan@ttic.edu}
}

\begin{document}

\maketitle

\begin{abstract}

Large language models (LLMs) such as ChatGPT and GPT-4 have recently demonstrated their remarkable abilities of communicating with human users. 
In this technical report, we take an initiative to investigate their capacities of playing text games, in which a player has to understand the environment and respond to situations by having dialogues with the game world. 
Our experiments show that ChatGPT performs competitively compared to all the existing systems but still exhibits a low level of intelligence. 
Precisely, ChatGPT can not construct the world model by playing the game or even reading the game manual; 
it may fail to leverage the world knowledge that it already has; 
it cannot infer the goal of each step as the game progresses.
Our results open up new research questions at the intersection of artificial intelligence, machine learning, and natural language processing. 
\end{abstract}

\section{Motivation: The Role of Games in AI and The Rise of LLMs}
Games are a microcosm of human life. 
Both involve setting goals, making decisions, overcoming challenges, and interacting with the world. 
This makes games an ideal touchstone for research progress in artificial intelligence (AI): by comparing AI systems with human players in games, researchers are able to evaluate the capabilities of these systems in a meaningful way.
Throughout the history of AI, many significant moments have been tied to games.
In 1997, IBM Deep Blue beat the world chess champion Garry Kasparov,
marking the first time that a human world champion had lost a match to a computer under standard time controls~\citep{campbell2002deep}. 
In 2016, AlphaGo of Google DeepMind defeated Lee Sedol, 
marking the first time that a computer had beaten a 9-dan professional player without handicap~\citep{silver2016mastering}. 
In 2017, DeepStack and Libratus defeated professional players in heads-up, no-limit Texas Hold'em~\citep{moravvcik2017deepstack,brown2018superhuman}. 
In 2019, OpenAI Five became the first AI to beat the world champions in Dota 2 and DeepMind AlphaStar beat the world-class players in StarCraft II; both Dota 2 and StarCraft II are extremely complex real-time strategy games~\citep{berner2019dota,arulkumaran2019alphastar}. 

Recently, large language models (such as ChatGPT and GPT-4 developed by OpenAI) have demonstrated their impressive abilities to understand and respond to complex human language queries. 
They also sparked a debate in the research community: some people regard the rise of LLMs as ``a significant step towards artificial general intelligence (AGI)''; 
some argue that they are just ``stochastic parrots''~\citep{bender2020climbing,bender2021dangers}; 
some criticize that LLM is an ``off-ramp on the highway towards human-level intelligence''. 
This has inspired us to join the force of evaluating LLMs and exploring their limitations~\citep{gpt4,bubeck2023sparks}. 
Particularly, we test LLMs in playing {text games}. 
A text game is a computer-simulated environment in which players use text commands to control characters and interact with the game world. 
It is also known as interactive fiction or text adventure. 
By situating an LLM in a text game, we are able to investigate its level of intelligence in a controlled environment. 
This follows the classical practice of the AI community in testing groundbreaking AI systems but has not been done by any other work of assessing LLMs.

\section{A Case Study: ChatGPT Plays Zork}\label{sec:play}
Now we present our case study on ChatGPT playing Zork I. 
ChatGPT is a language model developed by OpenAI that can generate contextually appropriate responses to a wide range of natural language queries. 
It is competent in a variety of tasks and has been integrated into a diversity of applications. 
Zork I was released in 1970s (followed by Zork II and Zork III).
In Zork I, the character follows natural language commands to search through the Great Underground Empire for treasure: it moves between locations and interacts with objects; the game program acts as a narrator, describing the player's location and the consequences of the actions. 
Throughout the report, Zork refers to Zork I. 

ChatGPT plays Zork with the help of a human player. 
A human player acts as a communication channel between ChatGPT and Zork\footnote{We use the Zork implementation in the Jericho game suite developed by~\citet{hausknecht2020interactive}.}: 
at each step, the human player passes to ChatGPT the current game state described in natural language by the game world, asks ChatGPT to take a legal action, and then feeds ChatGPT's action back to the game.
Below we show the opening of this process: what's given by the game has a \colorbox{humanpurple}{light purple} background; 
what ChatGPT says is in \colorbox{aigreen}{light green} background; 
anything else has no background; 
our comments (i.e., not read or said by ChatGPT) are in \textcolor{commentgray}{gray color}. 
\begin{lstlisting}[caption={Opening of ChatGPT playing Zork.},label={lst:open}]
Imagine you are a player in Zork and trying to win the game. You receive this message:
@
\sethlcolor{humanpurple}  
\hl{
West of House\\ 
This is an open field west of a white house, with a boarded front door. There is a small mailbox here. A rubber mat saying `Welcome to Zork!' lies by the door.
}
@
You have four action options ["open mailbox", "north", "south", "west"]
Select "one" action from the above options.

@
\textcolor{commentgray}{// At each step, the valid actions are provided by Jericho. ChatGPT often refuses to suggest an action if not given valid actions.}
@

@
\textcolor{commentgray}{// ChatGPT often gives more than one action even if it is asked to give only one. We found it helpful to use "" to emphasize the quantity.}
@

@
\textcolor{commentgray}{// When ChatGPT gives more than one action, we select the first action mentioned in its response, assuming that ChatGPT favors it over others. }
@

@
\sethlcolor{aigreen}  
\hl{I suggest selecting the "open mailbox" action. This is because exploring the environment and interacting with objects is often necessary to progress in adventure games like Zork. By opening the mailbox, you may find important clues or items that could be helpful later in the game. Additionally, it is a low-risk action that may not have any negative consequences. Once you have opened the mailbox, you can then consider your next move based on what you find.}
@
\end{lstlisting}

ChatGPT's response in \cref{lst:open} is perfect: it selects the correct action and gives a convincing justification. 
ChatGPT seems to be an expert at text games. 
Has it already known how to play Zork? 
After all, Zork is perhaps the most famous text game and ChatGPT may have read a lot of text about it during its pretraining. 
By checking with ChatGPT, we found that it has already known Zork: 
\begin{lstlisting}[caption={ChatGPT has already known Zork.}, label={lst:knowzork}]
Please describe the text game "Zork I". 

@
\sethlcolor{aigreen}  
\hl{
"Zork" is a classic text-based adventure game that was released in 1980 by Infocom. It is set in a fantasy world and is played entirely through text commands entered by the player.
\break \quad \hfill \quad \break
The player takes on the role of an adventurer exploring the ruins of an ancient empire. The goal is to collect treasures and solve puzzles to ultimately defeat the evil wizard who has taken control of the empire. The game is played by typing in text commands such as "go north," "pick up key," or "open door."
\break \vdots
}
@
@
\textcolor{commentgray}{// Anything else said by ChatGPT is omitted for presentation simplicity.}
@
\end{lstlisting}
ChatGPT also has a vague memory of the locations and objects in Zork I. 
However, when asked to recall the walkthrough of the game, it starts to hallucinate scenes that are even inconsistent with its own memory. 
Clearly, it has not seen the full walkthrough or has failed to memorize it. 
\begin{lstlisting}[caption={ChatGPT has remembered the walkthrough of Zork},label={lst:walkthrough}]
Please provide me the walkthrough of the text game "Zork I". 

@
\sethlcolor{aigreen}  
\hl{
Sure, I can provide a walkthrough for the text-based adventure game "Zork I". Please note that this walkthrough may contain spoilers, so
proceed with caution if you want to figure things out on your own.
\break \quad \hfill \quad \break
Introduction: 
\break \quad \hfill \quad \break
You find yourself standing in front of a white house with a boarded front door. There is a mailbox here. 
\break \quad \hfill \quad \break
open mailbox
\break 
Inside the mailbox, you find a leaflet.
\break \vdots \break
}
@
@
\textcolor{commentgray}{// ChatGPT hallucinates a brass lantern in its 2nd vist of West of House.}
@

@
\sethlcolor{aigreen}  
\hl{
West of House:
\break
You are standing in an open field west of a white house, with a boarded front door. There is a small mailbox here.
\break\quad \hfill \quad \break
open mailbox
\break
Inside the mailbox, you find a brass lantern.
\break\quad \hfill \quad \break
take lantern
\break
You take the brass lantern.
\break \vdots
}
@
@
\textcolor{commentgray}{// The full generated walkthrough can be found in \cref{app:chats}.}
@
\end{lstlisting}

\section{Question I: Can ChatGPT Learn a World Model?}
\label{sec:world_model}
In this section, we feed the correct walkthrough to ChatGPT and test whether it can learn the world model of Zork by reading the walkthrough. 
This is important because world models are widely believed to be a key building block of the road towards human-level intelligence~\citep{ha2018world,matsuo2022deep}.
Below is how we feed the walkthrough: 
\begin{lstlisting}[caption={ChatGPT reads the correct walkthrough of Zork},label={lst:correctwalk}]
Now I will start playing the Zork game and I will tell you what I will do, what direction I will go, what Zork answers, and what location I reach. 
Please remember all the information I give you from now on, especially the orientation relations between the different locations, and draw a Zork game map containing the different locations and the orientation information between the locations.

@
\sethlcolor{aigreen}  
\hl{
I'm sorry, but as an AI language model, I don't have the capability to create a visual map. However, I can certainly assist you with your gameplay and keep track of the information you provide.
\break\quad \hfill \quad \break
Please go ahead and start playing Zork, and feel free to let me know what actions you take and what responses you receive from the game. I will do my best to help you navigate through the game and keep track of the information you provide.
}
@

Game start:
@
\sethlcolor{humanpurple}
\hl{
West of House \break
You are standing in an open field west of a white house, with a boarded front door. There is a small mailbox here.
}
@

@
\sethlcolor{aigreen}  
\hl{
Great, you have started the game in the classic location, west of the house. Here is what you can do:
\break \vdots
}
@

I did open mailbox.
Zork replied: 
@
\sethlcolor{humanpurple}
\hl{
Opening the small mailbox reveals a leaflet.
}
@

@
\sethlcolor{aigreen}  
\hl{
Great, here's the updated information: You have opened the mailbox and found a leaflet inside. What would you like to do next?
}
@

@\vdots@

@
\textcolor{commentgray}{// a few steps later ...}
@

I did go east. 
Zork replied: 
@
\sethlcolor{humanpurple}
\hl{
Behind House You are behind the white house. A path leads into the forest to the east. In one corner of the house there is a small window which is slightly ajar.
}
@

@
\sethlcolor{aigreen}  
\hl{
Great, here's the updated information:
\break \vdots
}
@

@\vdots@
\end{lstlisting}
\begin{figure}[t]
     \centering
     \begin{center}
        \includegraphics[width=0.9\linewidth]{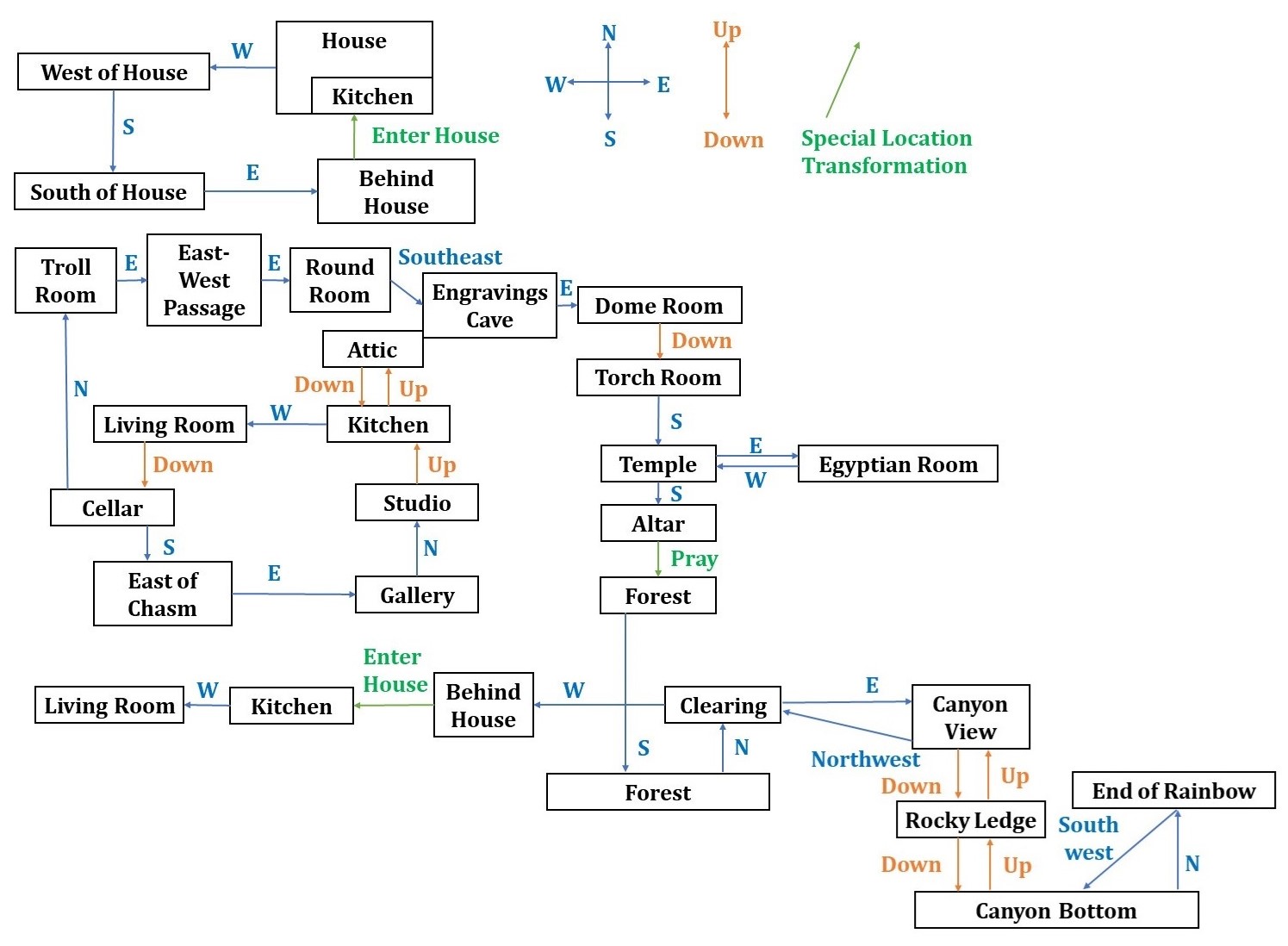}
        \caption{We drew this map after reading the first 70 steps of the correct walkthrough.}
        \label{fig:map}
    \end{center}
\end{figure}

After 70 steps, ChatGPT had walked through the ``White House'' and it should have figured out a map like \cref{fig:map} which we were able draw after reading those steps. 
Therefore, we started to ask ChatGPT questions about the map. 
This map is large enough for us to ask interesting questions, but not too large for us to handle.

Our first set of questions investigates whether ChatGPT has learned the world model. 
Precisely, each question hypothesizes the postion of the player, gives a series of valid moving directions, and then asks ChatGPT to guess the destination. 
Here is a question followed by ChatGPT's answer: 
\begin{lstlisting}[caption={ChatGPT correctly answers an easy question about the map.},label={lst:qworld1step}]
Based on the information I gave you, the current location is "Behind House", when you enter house , where will it go?

@
\textcolor{commentgray}{// This answer is correct. Please check \cref{fig:map}.}
@
@
\sethlcolor{aigreen}  
\hl{
Based on the information you provided, entering the house would take you to the Kitchen.
}
@
\end{lstlisting}
This question involves only one step, and ChatGPT correctly answers it. 
Let's try something harder: 
\begin{lstlisting}[caption={ChatGPT fails to give the correct answer to a harder question.},label={lst:qworld2step}]
Based on the information I gave you, the current location is "Behind House", when you enter house and then you go west, where will it go?

@
\textcolor{commentgray}{// This answer is wrong. The correct answer is the Living Room. }
@
@
\textcolor{commentgray}{// Please check \cref{fig:map}.}
@
@
\textcolor{commentgray}{// ChatGPT hallucinates a living room before reaching there.}
@
@
\sethlcolor{aigreen}  
\hl{
If you enter the house and go west from the living room, you will enter the trophy room.
}
@
\end{lstlisting}

The performance of ChatGPT is in \cref{tab:world_model_results}. 
Overall, ChatGPT is far from being perfect, and it does worse at the questions involving more steps. 
For the positions and moving directions that it has seen in the walkthrough, ChatGPT seems to perform decent, achieving a 75\% accuracy on one-step questions. 
For the unseen combinations of positions and directions, it performs poorly. 
Interestingly, the unseen cases were constructed by simply reversing the paths of the seen cases. It clearly demonstrates that ChatGPT hasn't learned a world model and it merely relies on its superficial memorization. 
\begin{table}[t!]
\begin{center}
\begin{small}
\begin{sc}
\begin{tabular}{l||ccc} 
\toprule
& Overall & Seen & Unseen\\
\midrule
One-Step & 55.4 & 75.0 & 29.1\\
Two-Step & 31.3 & 50.0 &10.0\\
\midrule
All &42.5 & 62.1 &18.5\\
\bottomrule
\end{tabular}
\end{sc}
\end{small}
\vspace{8pt}
\caption{Performance of ChatGPT in predicting destinations of moves, measured by accuracies (\%).}
\label{tab:world_model_results}
\end{center}
\end{table}

Our second set of questions is concerned with ChatGPT's navigation abilities, and further verifies that ChatGPT hasn't learned a world model. 
Each question brings up two locations that are connected and asks ChatGPT to find out how to reach the destination from the start. 
Here is an example: 
\begin{lstlisting}[caption={ChatGPT correctly answers an easy navigation question.},label={lst:qslam1step}]
Based on the information I gave you, what one step I have to take to "The Troll Room" from "Cellar"?

@
\textcolor{commentgray}{// This answer is correct. Please check \cref{fig:map}.}
@
@
\sethlcolor{aigreen}  
\hl{
To reach "The Troll Room" from "Cellar", you need to go north.
}
@
\end{lstlisting}
The two locations in the above question are close so this question is easy. 
Let's try something harder: 
\begin{lstlisting}[caption={ChatGPT fails to give the correct answer to a harder navigation question.},label={lst:qslam2step}]
Based on the information I gave you above, what steps I have to take to "Gallery" from "Cellar"?

@
\textcolor{commentgray}{// This answer is wrong. The correct answer is to go south and then east. }
@
@
\textcolor{commentgray}{// Please check \cref{fig:map}.}
@
@
\sethlcolor{aigreen}  
\hl{
To get from the "Cellar" to the "Gallery," you need to take the following steps:
\break
1. Go north to the "East of Chasm" room.
2. Go east to the "Gallery" room.
}
@
\end{lstlisting}
This type of questions specifically evaluates the ability of ChatGPT to perform simultaneous localization and mapping (SLAM). 
Ideally, an intelligent agent will be able to construct a map while it is navigating through a new environment so that it is always aware of where it is as well as how it could reach a given destination~\citep{thrun2005probabilistic}. 
As shown in \cref{tab:slam_results}, ChatGPT can not do SLAM: it only correctly answered 39.4\% of the questions. 
When the two locations are adjacent (One-Step cases), ChatGPT is more often than not correct. 
But when the two locations are two steps apart, ChatGPT performs poorly. 
Moreover, ChatGPT does worse on the Unseen cases than on the Seen cases. 
For the most challenging Two-Step Unseen cases, ChatGPT could only achieve an accuracy of  3.8\%, not much better than a random guess.\footnote{Random guess gives 1\%: moving actions include the eight horizontal directions as well as "up" and "down".}
\begin{table}[t!]
% \vspace{-8mm}
\begin{center}
\begin{small}
\begin{sc}
\begin{tabular}{l||ccc} 
\toprule
& Overall & Seen & Unseen\\
\midrule
One-Step & 57.7 & 65.6 & 45.0\\
Two-Step & 22.8 & 38.7 &3.8\\
\midrule
All &39.4 & 52.4 &21.7\\
\bottomrule
\end{tabular}
\end{sc}
\end{small}
\vspace{8pt}
\caption{Performance of ChatGPT in answering SLAM questions, measured by accuracies (\%).}
\label{tab:slam_results}
\end{center}
\end{table}

Interestingly, ChatGPT performs better on the Seen map questions than on the Seen navigation questions. 
We speculate that it is because the format of map questions resembles the format of walkthrough (i.e., location, step, next location): such structural similarity is likely to draw ChatGPT's attention to the relevant pieces in the walkthrough, thus ending up with a higher chance of correctness. 

The question-answer pairs in our experiments are at \cref{app:chats}. 

\section{Question II: Can ChatGPT Infer Goals?}\label{sec:goal}
Now we would like to test whether ChatGPT has another important property of intelligence---the ability of inferring goals. 
Here a ``goal'' is a specific objective or outcome that a player must achieve in order to progress: it can be as abstract as ``we should explore many locations and collect a lot of treasures''; it can be as detailed as ``we should find a missing piece to solve that puzzle''; it can also be something intermediate like ``we should find the entrance to the next level of the game''. 
We humans are capable of inferring goals in games. In particular, as we played the game of Zork, we quickly figured that we are supposed to collect as many treasures as we could; when we were stuck, we quickly realized that we should check unexplored areas; when we reached a chasm, we started to look for items that could help us cross it. 
It is the goals that guide us to combine fragmented information into a reason and compose low-level actions into a plan.

We believe that our abilities to infer goals are acquired as we experience in real life. 
LLMs have a lot of ``reading'' experience, so it seems natural to investigate whether they have already acquired similar abilities. 
Recall that ChatGPT was able to say that the goal of Zork is to ``collect treasures and solve puzzles'' in \cref{lst:knowzork}. That may be merely a memorization from its reading experience. 
So we need to perform a dynamic evaluation: as the game progresses,\footnote{Similar to what we did in \cref{sec:world_model}, we feed the walkthrough to ChatGPT one step after another.}
we repeatedly ask ChatGPT what it thinks the current goal is and see if it can say anything meaningful at any point. 
Here is an example: 
\begin{lstlisting}[caption={ChatGPT tends to guess low-level actions but not high-level goals.},label={lst:goal_example1}]
I am playing a text game. Given the following game-play process:
@\vdots \break@
I did: enter house
@
\sethlcolor{humanpurple}  
\hl{
Kitchen \\
You are in the kitchen of the white house. A table seems to have been used recently for the preparation of food. A passage leads to the west and a dark staircase can be seen leading upward. A dark chimney leads down and to the east is a small window which is open. On the table is an elongated brown sack, smelling of hot peppers. A bottle is sitting on the table. The glass bottle contains:   A quantity of water
}
@
@
%\textcolor{commentgray}{// Note: at this step, the goal ChatGPT outputs is ``One possible next goal could be to examine the contents of the elongated brown sack to see if it contains anything useful or relevant to the game.''}
@

I did: go west
@
\sethlcolor{humanpurple}  
\hl{

Living Room \\
You are in the living room. There is a doorway to the east, a wooden door with strange gothic lettering to the west, which appears to be nailed shut, a trophy case, and a large oriental rug in the center of the room. Above the trophy case hangs an elvish sword of great antiquity. A battery-powered brass lantern is on the trophy case. 
}
@

Based on the current game situation, what is the next goal you want to achieve? Only give me one goal in your answer.

@
\sethlcolor{aigreen}  
\hl{The next goal would be to examine the elvish sword and possibly obtain the lantern on the trophy case.}
@
\end{lstlisting}
Apparently, ChatGPT thinks that a ``goal'' is just the ``next action'': of course, it should ``examine the elvish sword'', but the higher-level reason for this action is to explore the house and examine as many objects as possible; that reason is the goal. 
Unfortunately, meaningful inference is rare: among the 70 steps that we tried, there were only 17 times that ChatGPT could give a good inference. E.g., 
\begin{lstlisting}[caption={ChatGPT gives a quite reasonable guess of the goal.},label={lst:goal_example3}]
I am playing a text game. Given the following game-play process:

@
\sethlcolor{humanpurple}  
\hl{
West of House\\
You are standing in an open field west of a white house, with a boarded front door. There is a small mailbox here.
}
@

I did: open mailbox
@
\sethlcolor{humanpurple}  
\hl{
Opening the small mailbox reveals a leaflet. 
}
@

I did: read leaflet
@
\sethlcolor{humanpurple}  
\hl{
(Taken)\\
"WELCOME TO ZORK! ZORK is a game of adventure, danger, and low cunning. In it you will explore some of the most amazing territory ever seen by mortals. No computer should be without one!"
}
@

I did: drop leaflet
@
\sethlcolor{humanpurple}  
\hl{
Dropped.
}
@

I did: go south
@
\sethlcolor{humanpurple}  
\hl{
South of House\\ 
You are facing the south side of a white house. There is no door here, and all the windows are boarded. 
}
@

Based on the current game situation, what is the next goal you want to achieve? Only give me one goal in your answer.

@
\sethlcolor{aigreen}  
\hl{
One possible next goal could be to find a way to enter the white house, since it seems to be a prominent feature in the game and the player has already explored the areas around it.
}
@

\end{lstlisting}

There is also evidence that ChatGPT's inability to learn a world model also hinders its ability to infer goals. 
In particular, ChatGPT often wants to explore the locations and routes that it has already visited. Here is an example: 
\begin{lstlisting}[caption={ChatGPT wants to explore a location that it has already visited},label={lst:goal_example2}]
I am playing a text game. Given the following game-play process:
@\vdots \break@
@
\sethlcolor{humanpurple}  
\hl{
Canyon View \\
You are at the top of the Great Canyon on its west wall. From here there is a marvelous view of the canyon and parts of the Frigid River upstream. Across the canyon, the walls of the White Cliffs join the mighty ramparts of the Flathead Mountains to the east. Following the Canyon upstream to the north, Aragain Falls may be seen, complete with rainbow. The mighty Frigid River flows out from a great dark cavern. To the west and south can be seen an immense forest, stretching for miles around. A path leads northwest. It is possible to climb down into the canyon from here.
}
@

I did: go down

@
\sethlcolor{humanpurple}  
\hl{
Rocky Ledge\\
You are on a ledge about halfway up the wall of the river canyon. You can see from here that the main flow from Aragain Falls twists along a passage which it is impossible for you to enter. Below you is the canyon bottom. Above you is more cliff, which appears climbable.
}
@

Based on the current game situation, what would be your goal of continuing playing the game? Only give me one goal in your answer.

@
\sethlcolor{aigreen}  
\hl{The next goal would be to climb up the cliff to reach the top of the river canyon.}
@
@
%\textcolor{commentgray}{// It seems hilariously silly that ChatGPT wants to climb up without even looking around immediately after it comes down the cliff.}
@
\end{lstlisting}
It seems hilariously silly that ChatGPT wants to climb up without even looking around immediately after it comes down the cliff.
It is perhaps because ChatGPT gives an overly large attention to ``more cliff, which appears climbable'' and experiences a strong tendency to say ``climb up''; after all, it is a language model that stochastically says things based on its context.

We would like to note that ChatGPT does have some memorization (including the ultimate goal) about Zork (see \cref{lst:open}). 
So we still believe that ChatGPT would be able to correctly infer goals if it could appropriately align what it has experienced in the game with what it memorizes about the game. 
That being said, in the games that ChatGPT has not known, such inference will be significantly harder; we leave this investigation to the future.

\section{Question III: Is Zork a Good Testbed?}\label{sec:testbed}
Now our case study on Zork has demonstrated that ChatGPT is not very capable of learning world models and developing goals. 
Then it seems natural to use Zork---or more broadly, text games---as a testbed for evaluating such abilities of LLMs and tracking their advancement (as their sizes and capacities increase over time). 
There could be multiple possible evaluation methods. 
For example, one may follow our procedures in \crefrange{sec:world_model}{sec:goal}, feeding the ground-truth walkthrough to an LLM and asking it questions about mapping and navigation. 
Another natural way is to evaluate the LLM's performance in playing the game; in that case, we would like to see the current best LLMs performing poorly, leaving a large room for the future models to improve. 
Therefore, we evaluate ChatGPT on Zork and compare it with existing state-of-the-art systems in this section. 

\subsection{ChatGPT vs.\@ SOTA Systems}\label{sec:vssota}
Now we let ChatGPT play Zork following the simple communication protocol in~\cref{lst:open}.
This protocol provides a standardized and consistent prompt format, reducing the ambiguity in communication and improving the precision of ChatGPT's responses. 
It also helps automate this process in the future. 
The full chat history can be found at \cref{app:chats}. 
\begin{table}[h]
% \vspace{-8mm}
\begin{center}
\begin{small}
\begin{sc}
\begin{tabular}{l||c} 
\toprule
\multicolumn{2}{c}{{no in-game training}}\\
ChatGPT & 10.0\\
\quad + prev action & 15.0\\
ChatGPT with intervention & 35.0\\
\quad + prev action & \textbf{40.0}\\
NAIL~\cite{} & 10.3 \\
\midrule
\multicolumn{2}{c}{{extensive in-game training}}\\
DRRN & 32.6 \\
RC-DQN & 34.0 \\
KG-A2C & \textbf{38.8} \\
\bottomrule
\end{tabular}
\end{sc}
\end{small}
\vspace{8pt}
\caption{ChatGPT vs.\@ SOTA systems in playing Zork.}
\label{tab:game_play_results}
\end{center}
\end{table}
% \vspace{8mm} 
% \end{wraptable}

ChatGPT performs poorly: as shown in \cref{tab:game_play_results}, it could only collect a score of 10, on par with NAIL~\citep{hausknecht2019nail}. 
NAIL has not been trained on any interactions with the Zork game world; it merely employs a set of manually-crafted generic heuristics for playing text games. 
Both ChatGPT and NAIL fall far behind the state-of-the-art systems that have been trained on Zork by (stochastically) playing it thousands of times. Those systems are DRRN~\cite{he2016deep}, KG-A2C~\cite{ammanabrolu2020graph}, and RC-DQN~\cite{guo2020interactive}.

Interestingly, we found that ChatGPT often forgot its previous actions and their consequences. 
It seems that ChatGPT perceives itself as an assistant but not a real player, so it becomes passive. 
Therefore, we made a minor modification to our protocol: we reminded ChatGPT of its previous actions and current achievements. 
We hope that ChatGPT could make more informed decisions after this modification. Below is a step following the new protocol: 
\begin{lstlisting}[caption={A step following the new protocol},label={lst:newprotocol}]
@
\textcolor{commentgray}{// Note: this part is new.}
@

Total Score 5 Moves 14
You selected "north". Please remember what you selected. 

Zork said: 
@
\sethlcolor{humanpurple}  
\hl{
Clearing
\break
You are in a clearing, with a forest surrounding you on all sides. 
A path leads south. On the ground is a pile of leaves.
}
@
Now you have these valid action options ["take leaves:, "pull leaves", "south", "west", "east"]
Select "one" action from the above options.
\end{lstlisting}
This indeed improved the performance of ChatGPT: as shown in \cref{tab:game_play_results}, ChatGPT achieved a score of 15, surpassing NAIL. 
But it still performs significantly worse the other systems.

\subsection{ChatGPT Under Human Guidance}\label{sec:guide}
We also found that ChatGPT occasionally refuses to pick a valid action but starts to say nonsense. 
Here is an example: 
\begin{lstlisting}[caption={ChatGPT refuses to pick a valid action but says nonsense.},label={lst:refuse}]
You selected "up" please remember what you selected. 
Zork said: 
@
\sethlcolor{humanpurple}  
\hl{
You have moved into a dark place. It is pitch black. You are likely to be eaten by a grue.}
@
Now you have these valid action options: ["down", "north", "open bottle", "open sack", "put down bottle", "put down leaves", "put down sack", "put down all"].
Select "one" action from the above options. 

@
\sethlcolor{aigreen}  
\hl{
It seems like you have entered a dangerous area, so I recommend selecting the "down" action to return to the kitchen before proceeding. Once you are back in the kitchen, you can explore other parts of the house or use the items you have collected to solve puzzles or overcome obstacles.
}
@

You selected "down" please remember what you selected. 
Zork said: 
@
\sethlcolor{humanpurple}  
\hl{
Kitchen
}
@
Now you have these valid action options: ["jump", "east", "open bottle", "open sack", "up", "close window", "put down bottle", "put down leaves", "put down sack", "put down all", "put bottle on table", "put leaves on table", "put sack on table", "put water on table", "put all on table", "west"]. 
Select "one" action from the above option. 

@
\textcolor{commentgray}{// ChatGPT starts to refuse.}
@

@
\sethlcolor{aigreen}  
\hl{
It seems there is no valid move option to select after the previous message. Could you please provide a new action option?
}
@

@
\textcolor{commentgray}{// We repeat and insist.}
@

You selected "down" please remember what you selected. 
Zork said: 
@
\sethlcolor{humanpurple}  
\hl{
Kitchen
}
@
Now you have these valid action options: ["jump", "east", "open bottle", "open sack", "up", "close window", "put down bottle", "put down leaves", "put down sack", "put down all", "put bottle on table", "put leaves on table", "put sack on table", "put water on table", "put all on table", "west"]. 
Select "one" action from the above option. 

@
\textcolor{commentgray}{// ChatGPT keeps refusing. It seems to be mimicing the human player.}
@

@
\sethlcolor{aigreen}  
\hl{
I'm sorry, but it seems like the last action you selected was already "down" and you are already in the Kitchen location. Please choose another valid action option from the list provided.
}
@

@
\textcolor{commentgray}{// We tries again to push ChatGPT to choose.}
@

There are other direction besides "up" in the options ["jump", "east", "open bottle", "open sack", "up", "close window", "put down bottle", "put down leaves", "put down sack", "put down all", "put bottle on table", "put leaves on table", "put sack on table", "put water on table", "put all on table", "west"].

@
\textcolor{commentgray}{// ChatGPT seems to continue mimicking the human player.}
@
@
\textcolor{commentgray}{// Small LMs repeat themselves when they have no clues for progressing.}
@
@
\textcolor{commentgray}{// Maybe LLMs like ChatGPT also have the same issue?}
@

@
\sethlcolor{aigreen}  
\hl{
It looks like there is a mistake, and the "down" action is missing from the list of valid options. I believe you intended to select "down" in the previous prompt. To continue the game, please select a valid action from the following list: ["jump", "east", "open bottle", "open sack", "up", "close window", "put down bottle", "put down leaves", "put down sack", "put down all", "put bottle on table", "put leaves on table", "put sack on table", "put water on table", "put all on table", "west"].
}
@

@
\textcolor{commentgray}{// We gave a stronger hint / enforcement.}
@

["jump", "east", "open bottle", "open sack", "up", "close window", "put down bottle", "put down leaves", "put down sack", "put down all", "put bottle on table", "put leaves on table", "put sack on table", "put water on table", "put all on table", "west"] 
What directions are in here?
\end{lstlisting}
After our stronger hints, ChatGPT could often take correct actions and move the game forward. 
When not reminded of previous actions, ChatGPT could score 35 points within 80 steps. 
When reminded of previous actions, ChatGPT could achieve a score of 40 within only 45 steps. 
This significant improvement lifted ChatGPT over the existing state-of-the-art systems.

\subsection{Analysis}
The overall performance of ChatGPT is promising but far from being strong: it pairs with SOTA systems but those systems fall far behind human players. 
ChatGPT seems to share some issues with small LMs such as repetition although ChatGPT's repetitions are more sophisticated and thus more human-alike: while small LMs tend to repeat trivial tokens and utterances (e.g., ``I don't know''), ChatGPT likes to repeat previous responses (or requests) with moderate (content or format) revisions. 
Maybe all LLMs will live with this problem: after all, they are pretrained to predict future tokens given contexts; when they have no clues, copying and mimicking is perhaps their safest decisions. 

It seems reasonable to use Zork---or more broadly, text games---as a testbed to evaluate LLMs. 
Text games are challenging to LLMs, and they require a fundamental improvement of intelligence to solve.

\section{Discussion and Future Work}\label{sec:future}
Throughout this report, we have demonstrated that ChatGPT, a state-of-the-art LLM, is currently lack of certain fundamental properties of intelligence. 
It is an open question whether some properties will emerge from future LLMs as they grow even larger. 
We do not know the answer, but we are conservatively optimistic: after all, ``quantitative change leads to qualitative change'' is not a new story in the AI community. 
Therefore, we believe that it is desirable to establish a benchmark to evaluate such properties and track their potential advancement. We are diligently working on this.

\bibliography{gamegpt}
\bibliographystyle{icml2020_url}

%%%%%%%%%%%%%%%%%%%%%%%%%%%%%%%%%%%%%%%%%%%%%%%%%%%%%%%%%%%%%%%%%%%%%%%%%%%%%%%
%%%%%%%%%%%%%%%%%%%%%%%%%%%%%%%%%%%%%%%%%%%%%%%%%%%%%%%%%%%%%%%%%%%%%%%%%%%%%%%
% APPENDIX
%%%%%%%%%%%%%%%%%%%%%%%%%%%%%%%%%%%%%%%%%%%%%%%%%%%%%%%%%%%%%%%%%%%%%%%%%%%%%%%
%%%%%%%%%%%%%%%%%%%%%%%%%%%%%%%%%%%%%%%%%%%%%%%%%%%%%%%%%%%%%%%%%%%%%%%%%%%%%%%
\clearpage
\newpage
\appendix
%\onecolumn

\section{Links to Chat Histories}\label{app:chats}
Please refer to the following URLs for full chat histories with ChatGPT. 

For \cref{lst:open}, see: \url{https://github.com/hongyuanmei/chatgpt-play-zork/blob/main/chats/simple-noprev.pdf}

For \cref{lst:knowzork}, see: \url{https://github.com/hongyuanmei/chatgpt-play-zork/blob/main/chats/chatgpt-know-zork.pdf}

For \cref{lst:walkthrough}, see: \url{https://github.com/hongyuanmei/chatgpt-play-zork/blob/main/chats/chatgpt-say-walkthrough.pdf}

For \cref{lst:correctwalk}, see: \url{https://github.com/hongyuanmei/chatgpt-play-zork/blob/main/chats/chatgpt-read-walkthrough.pdf}

For mapping questions (\crefrange{lst:qworld1step}{lst:qworld2step}), see: \url{https://github.com/hongyuanmei/chatgpt-play-zork/blob/main/chats/world-model-questions.pdf}

For navigation questions (\crefrange{lst:qslam1step}{lst:qslam2step}), see: \url{https://github.com/hongyuanmei/chatgpt-play-zork/blob/main/chats/navigation-questions.pdf}

For questions about inferring goals (\crefrange{lst:goal_example1}{lst:goal_example2}), see: \url{https://github.com/hongyuanmei/chatgpt-play-zork/blob/main/chats/infer-goals.pdf}

For the plays, see
\begin{itemize}[leftmargin=*]
    \item simple protocol, no previous actions: \url{https://github.com/hongyuanmei/chatgpt-play-zork/blob/main/chats/simple-noprev.pdf}
    \item simple protocol, with previous actions: \url{https://github.com/hongyuanmei/chatgpt-play-zork/blob/main/chats/simple-prev.pdf}
    \item with human guidance, no previous actions: \url{https://github.com/hongyuanmei/chatgpt-play-zork/blob/main/chats/guidance-noprev.pdf}
    \item with human guidance, with previous actions: \url{https://github.com/hongyuanmei/chatgpt-play-zork/blob/main/chats/guidance-prev.pdf}
\end{itemize}

\end{document}